\documentclass[runningheads]{llncs}

\usepackage[T1]{fontenc}

\usepackage{csquotes}

\usepackage{graphicx}

\usepackage{hyperref}
\usepackage{color}

\usepackage{subcaption} 

\usepackage{pgfplots}
\usepackage{algorithm}
\usepackage{algorithmic}
\pgfplotsset{compat=1.18}

\usepackage{enumitem}
\usepackage{textcomp}
\usepackage{booktabs}
\usepackage{wrapfig}
\usepackage{array}
\usepackage{amssymb}
\usepackage[acronym]{glossaries}
\glsdisablehyper
\newacronym{tda}{TDA}{Topological Data Analysis}
\newacronym{sota}{SOTA}{State of the Art}
\newacronym{ph}{PH}{Persistent Homology}
\newacronym{pd}{PD}{Persistence Diagram}
\newacronym{htr}{HT}{Hough Transformation}
\newacronym{tht}{THT}{Topological Hough Transformation}

\usepackage{xspace}
\newcommand{\tda}{\gls{tda}\xspace}
\newcommand{\ph}{\gls{ph}\xspace}
\newcommand{\pd}{\gls{pd}\xspace}
\newcommand{\sota}{state-of-the-art\xspace}

\newcommand{\eg}{e.g.\xspace}
\newcommand{\ie}{i.e.\xspace}

\usepackage{color}
\usepackage{mathtools}
\usepackage{tcolorbox}
\tcbuselibrary{skins,breakable}
\usetikzlibrary{shadings,shadows}

\newcommand{\ignore}[1]{}

    {\endtcolorbox}

\usepackage{xcolor}

\definecolor{mblue}{HTML}{648FFF}
\definecolor{mpurple}{HTML}{785EF0}
\definecolor{mred}{HTML}{DC267F}
\definecolor{mgreen}{HTML}{3CDAAC}
\definecolor{morange}{HTML}{FE6100}
\definecolor{myellow}{HTML}{FFB000}
\usepackage{cleveref}
\usepackage{amsfonts}

\DeclareMathOperator{\id}{id}

\begin{document}

\title{Persistence-based Line Detection in Hough Space}
\begingroup
\renewcommand\thefootnote{}\footnotetext{This version of the article has been
  accepted for publication, after peer review, and is subject to Springer
  Nature's AM terms of use, but is not the Version of Record and does not
  reflect post-acceptance improvements or any corrections. The Version of
  Record of this article, first published in \emph{Data Science -- Analytics
  and Applications} (Springer, 2027, pp.~85--94), is available online at
  \url{https://doi.org/10.1007/978-3-032-13233-8_9}.}
\addtocounter{footnote}{-1}
\endgroup

\titlerunning{Persistence-based Line Detection in Hough Space}

\newboolean{anonymized}

\setboolean{anonymized}{false}

\ifthenelse{\boolean{anonymized}}
{
    \author{First Author\inst{1} \and
    Second Author\inst{1,2} \and
    Third Author\inst{1,2} \and
    Fourth Author\inst{1,2} \and
    Fifth Author\inst{1,2}
    }

    \authorrunning{F.~Author et al.}

    \institute{Department for Double Blind Review,
    \and Research Center for Double Blind Review Studies,\\Anon Universisty, City, Country\\
    \email{\{ first.author, second.author, third.author, fourth.author, fifth.author \}@anon-university.com}}
}{
    \author{
        Johannes Ferner\inst{1}\thanks{Authors listed in alphabetical order. J.~Ferner and A.~Pop contributed
            equally to this work. M.~Uray is the corresponding author.
        }
        \and Stefan Huber\inst{1,2}{\textsuperscript{\thefootnote}}
        \and Saverio Messineo\inst{1,2}{\textsuperscript{\thefootnote}}
        \and Angel Pop\inst{1,2}{\textsuperscript{\thefootnote}}
        \and Martin Uray\inst{1,2}{\textsuperscript{\thefootnote}}
    }

    \authorrunning{J.~Ferner et al.}

    \institute{
        Department for Information Technologies and Digitalisation,
        \and Josef Ressel Centre for Intelligent
        and Secure Industrial Automation,\\Salzburg University of Applied Sciences, Puch bei Hallein, Austria\\
        \email\{j.ferner.itsb-m2023, stefan.huber, saverio.messineo, angel-ioan.pop,  martin.uray\}@fh-salzburg.ac.at}
}

\maketitle              

\begin{abstract}
    The Hough transform is a popular and classical technique in computer vision
    for the detection of lines (or more general objects). It maps a pixel into
    a dual space -- the Hough space: each pixel is mapped to the set of lines
    through this pixel, which forms a curve in Hough space.

    The detection of lines then becomes a \emph{voting} process to find those
    lines that received many votes by pixels. However, this voting is done by
    thresholding, which is susceptible to noise and other artifacts.

    In this work, we present an alternative voting technique to detect peaks in
    the Hough space based on persistent homology, which very naturally
    addresses limitations of simple thresholding. Experiments on synthetic data
    show that our method significantly outperforms the original method,
    while also demonstrating enhanced robustness.

    This work seeks to inspire future research in two key directions. First, we
    highlight the untapped potential of Topological Data Analysis techniques
    and advocate for their broader integration into existing methods, including
    well-established ones.
    
    Secondly, we initiate a discussion on the mathematical stability of the
    Hough transform, encouraging exploration of mathematically grounded
    improvements to enhance its robustness.

\end{abstract}

\section{Introduction}\label{sec:introduction}

Line detection is a fundamental task in computer vision and pattern recognition,
and the \emph{Hough transform} is a classical and most widely used technique for it~\cite{mukhopadhyay2015}.
It maps sampled points from the image space into a dual representation known
as the Hough space or parameter space, which is the space of lines, where each point is mapped to the set of all possible lines passing through it.
The line detection is then performed by applying a threshold to vote counts in this space~\cite{duda1972}, i.e., to find the lines that were voted most by points.
Existing \sota methods typically either rely on problem-dependent thresholds that require tuning or on computational
expensive approaches, such as machine learning, to identify \emph{relevant}
lines.

A key limitation of the Hough transform is its sensitivity to threshold selection,
with no universal value due to factors like noise, parallel line proximity, and
straightness deviation.
Various heuristics and multi-parameter methods have been proposed to address this issue,
but they inevitably introduce further parameters that require tuning,
making them equally sensitive to varying conditions.

In this paper we revisit the core problem: identifying lines within the Hough
space by finding the \enquote{relevant} peaks in the accumulation of votes in
the Hough space. This accumulation can be seen as a landscape, a scalar field,
in which we aim to identify the most \enquote{prominent} peaks.

However, the most prominent peaks are not necessarily the highest ones, as
found by thresholding. This is analogous to alpinism, where a large rock close
to a mountain peak may not be considered a mountain on its own. To make this
notion of \enquote{prominence} precise we use the mathematical framework of
\ph, which has already used in other domains for robust peak detection,
cf.\xspace~\cite{Huber2020}.

This work makes the following key contributions:
\begin{enumerate}[label=\alph*.)]

    \item We introduce a novel approach that integrates \ph into the widely
        used Hough transform, improving robustness against noise and reducing
        needs for parameter tuning.

    \item Within our experimental scenarios, our method empirically outperforms
        the original line detection method with respect to stability against
        noise and robustness against different number of samples for different
        lines.

    \item We experimentally provide evidence for \emph{Lipschitz stability} for
        the proposed method.

\end{enumerate}

In \Cref{sec:theory} we give an overview of the line detection using the Hough transform and
\ph, before describing related adaptions to the Hough
transform in Hough space in \Cref{sec:related-work}.
We continue by introducing our proposed method in \Cref{sec:method}, before
evaluating the proposed approach
in \Cref{sec:experiments-and-evaluation}.
The results, limitations and further potential are discussed in
\Cref{sec:discussion-conclusion}.

\section{Preliminaries}\label{sec:theory}
This section will introduce the theoretical background for line detection using the Hough transform
(\Cref{subsec:ht}) and \ph (\Cref{subsec:ph}).
In the following description, we limit the formal details to a level necessary
for the remaining work.
For further details, the reader is referred to the cited literature.

\subsection{Line Detection using the Hough Transform}\label{subsec:ht}

For the line detection using the Hough transform~\cite{duda1972}, we consider a
binarized input image \( P = \left\{ (x_i, y_i) \mid 1 \le i \le M \right\} \),
where $M$ is the number of points in the image, in which lines are to be
detected. A straight line can be parametrized by two parameters in different
ways. We use the orthogonal supporting vector in polar coordinates $(\rho,
\theta)$. The Hough space $S$ is then the $\rho$-$\theta$-plane; each point in it
describes a line in the image space. Note that each image point $(x, y)$ on a
line parametrized by $(\rho, \theta)$ fulfills $\rho = x \cos \theta + y \sin
\theta$. In other words, the set of lines through an image point $(x_i, y_i)$
forms in a sinusoidal curve $\rho = x_i \cos \theta + y_i \sin \theta$ in the
Hough space, cf.~\Cref{fig:hough-transform}.

If the image points lie on a straight line $(\rho, \theta)$ in image space $I$ then their
corresponding curves in Hough space intersect at a common point $\left( \rho,
\theta \right)$ in $S$. The principle idea can be generalized to other
geometric templates, such as circles, and can even be generalized to arbitrary
shapes~\cite{ballard1981}.

\begin{figure}
    \centering
    \begin{subfigure}[t]{0.38\linewidth}
    \centering
    \begin{tikzpicture}
        \begin{axis}[
            width=\linewidth,
            xlabel={$x$}, ylabel={$y$},
            axis lines=middle,
            xmin=0, xmax=9.0,
            ymin=0, ymax=9.0,
            grid=major
        ]
            % Points in Cartesian Space
            \addplot[only marks, mark=*, color=mred] coordinates {
                (2, 3)
            };
            \addplot[only marks, mark=*, color=mblue] coordinates {
                (4, 5)
            };
            \addplot[only marks, mark=*, color=mgreen] coordinates {
                (6, 7)
            };
            \node at (axis cs:2,3.5) [anchor=west] {$(x_1, y_1)$};
            \node at (axis cs:4,5.5) [anchor=west] {$(x_2, y_2)$};
            \node at (axis cs:6,7.5) [anchor=west] {$(x_3, y_3)$};
        \end{axis}
    \end{tikzpicture}
    \caption{Image Space $I$}
    \label{fig:cartesian-space}
\end{subfigure}
    ~
    \begin{subfigure}[t]{0.38\linewidth}
    \centering
    \begin{tikzpicture}
        \begin{axis}[
            width=\linewidth,
            xlabel={$\theta$},
            ylabel={$\rho$},
            axis lines=middle,
            xmin=0, xmax=3.14,
            ymin=-10.5, ymax=10.5,
            grid=major,
            clip mode=individual
        ]
            % Sinusoidal Curves in Hough Space
            \addplot[smooth, color=mred, thick] {2*cos(deg(x)) + 3*sin(deg(x))};
            \addplot[smooth, color=mblue, thick, dashed] {4*cos(deg(x)) + 5*sin(deg(x))};
            \addplot[smooth, color=mgreen, thick, dotted] {6*cos(deg(x)) + 7*sin(deg(x))};
        \end{axis}
    \end{tikzpicture}
    \caption{Hough Space $S$}
    \label{fig:hough-space}
\end{subfigure}
    \caption{The Hough transform maps image points (left)
    to sinusoidal curves in Hough space (right).}
    \label{fig:hough-transform}
\end{figure}

Note that the line $(\rho, \theta + \pi)$ equals the line $(-\rho, \theta)$. We
can denote this by the equivalence relation $(\rho, \theta + \pi) \equiv
(-\rho, \theta)$ in the Hough space $S$. Topologically speaking, we are
actually interested in the quotient space $S / \equiv$.

In practice this means we limit $S$ as the $\rho$-$\theta$-plane to $\mathbb{R}
\times [-\frac{\pi}{2}, \frac{\pi}{2})$ and obtain the topology of the Möbius strip.

The standard implementation of the line detection~\cite{duda1972} follows a two-step
process.
First, $S$ is discretized into an \emph{accumulator array} of size
$n_\rho \times n_\theta$, where $n_\rho, n_\theta \in \mathbb{R}$ are
application-dependent parameters.
Each cell in the accumulator array stores the number of functions passing
through it, effectively capturing the vote count for the corresponding line
parameters.
Second, the accumulator array is exhaustively searched for maximum values, where the corresponding positions
indicate parameters of the detected lines.

The Hough transform is widely applied in a various domains, including
facial feature detection, object recognition and tracking, and underwater
pipeline detection~\cite{hassanein2015}.
Its practical utility is reinforced by its implementation in
\emph{OpenCV}~\cite{opencv_library}, making it accessible for a range of
applications.
This implementation differs from the standard original approach, as it
does not search for the maximum value in the
\emph{accumulator array} to identify for a single specific line.
Instead, it applies a
threshold relative to the highest \emph{peak}, detecting all lines who exceed this predefined level.
This approach makes the method more flexible, allowing multiple lines to be
detected.

The performance of the Hough transform is limited by several factors, with its
high sensitivity to noise being a major concern.
For instance, if a few sampled point deviates from the exact position on the
line, the intersection in $S$ will spread, causing a reduction in \emph{votes}
and the appearance of a further peak.
As the number of noisy samples increases, multiple intersections accumulate,
eventually leading to false positive detections.

\subsection{Persistent Homology}\label{subsec:ph}

Topology is a branch of mathematics that studies
properties of \emph{objects} that remain unchanged under continuous
deformations.
At the intersection of data science, computer science and algebraic topology,

\glsfirst{tda} is an emerging field that analyzes the
topological and geometrical structures of datasets, capturing features such as connected
components, loops, and voids.

\ph~\cite{edelsbrunner2008,Huber2020} is a fundamental tool in \tda
that can be applied to a wide range of topological structures, including point
clouds~\cite{edelsbrunner2008} and
images~\cite{Huber2020}.
The key idea behind \ph is to describe the evolution of a dataset's topological features
as a parameter continuously varies.

This approach eliminates the need to select a fixed parameter while providing
deeper insights by introducing an additional dimension to the analysis.

The combinatorial structures that capture a dataset's topological features are so-called
\emph{complexes}.
To construct an evolving sequence of complexes from a topological structure $X$,
a varying parameter $r$ must be chosen\footnote{The application of
multiparameter persistence is not in the scope of this work.}.
For each $r \leq s$, we obtain a nested sequence of subcomplexes $ X_0 \subset \ldots X_r \subset X_s = X $, where $X_r$
is formed by applying the parameter $r$ on $X$.

The resulting sequence of nested subcomplexes is called \emph{filtration}.

\begin{definition}[Sublevelset Filtration]\label{theorem:superlevel-set}
    Let \( f: X \to \mathbb{R} \) be a continuous function on a topological
    space \( X \).
    The \emph{superlevel-set filtration} is the nested sequence of subspaces
    \(
        X_r = \{ x \in X \mid f(x) \geq r \}, \quad \forall r \in \mathbb{R}.
    \)
    This filtration satisfies
    \(
        X_s \subseteq X_r, \quad \forall s \leq r,
    \)
    and captures the evolution of topological features as \( r \) varies.
\end{definition}

For image analysis, we represent the input scalar fields as \emph{cubical complexes}, which are topological spaces constructed by joining unit cubes along their faces in a structured way.
On these cubical complexes we apply the
\emph{superlevel-set} filtration (\Cref{theorem:superlevel-set}).
Intuitively, the process of this particular filtration can be visualized as gradually draining water from a
flooded image.
As the water level decreases, distinct regions (``islands'') will emerge and progressively merge
until the entire underlying structure is revealed.

The resulting sequence of nested subcomplexes structure is transformed into a~\pd,
encoding the evolution of the data's homological features derived from the filtration.
In the \pd (\Cref{fig:ph-vs-ht} as an example), the topological features are represented based
on their \emph{birth} (the parameter value at which they first appear) on the x-axis
and their \emph{death} (the moment they merge with more prominent feature) on the y-axis.\footnote{
    The axis labels for both axes are reversed to ensure compatibility with the
    visualization of other filtrations.
}
All points lie above the diagonal, as no feature can die before it is born.
A features \emph{persistence}, defined as the difference between its death
and birth value, corresponds to its orthogonal distance from the diagonal.

\begin{remark}

    The~\pd can be visualized in various ways.
    For comparability with other types of filtration, we decided to plot the
    death over the birth level with decreasing values on both axes.
\end{remark}

Features with higher persistence are more significant, whereas those
closer to the diagonal are regarded as insignificant, \eg due to noise.
An intuitive analogy can be drawn from alpinism: persistent features resemble prominent mountain peaks.
The persistence of a peak is not solely determined by its absolute height
but rather how much it stands out relative to its surroundings.
An isolated peak remains highly prominent, whereas a tall peak situated near an
even taller one appears less significant.
Similarly, in persistent homology, features that persist over a long range of
the filtration parameter are considered more meaningful than those that quickly
vanish.

\section{Related Work}\label{sec:related-work}

Several adaptations of the Hough transform have been introduced to mitigate
noise-induced uncertainty within the Hough space $S$.
The Probabilistic Hough transform~\cite{stephens1991}, based on a maximum
likelihood assumption, models $S$ as a probabilistic space.
While this approach enhance robustness, it
increased computational complexity.

Similarly, the kernel-based Hough transform~\cite{fernandes2008} applies an
elliptical-Gaussian kernel to model uncertainty around detected clusters but also
introduces additional parameters, requiring
fine-tuning.

Another approach to analyzing the Hough space statistically involves the concept of
\emph{butterfly shapes}~\cite{du2012,xu2014}.
Instead of relying solely on peak detection within the accumulator array, these
methods consider the entire surface formed by the intersecting sinusoidal
functions, enabling for statistical analysis of peak parameters through
interpolation and variance analysis.

From an information-theoretic perspective, voting can be interpreted
as a random variable, with values modeled as probability
distributions~\cite{xu2014a,xu2015}.
Entropy measures derived from these distributions are further used to estimate peak
parameters, potentially enabling a more adaptive detection mechanism.

Additionally, machine learning-based approaches have been explored for line
detection, using methods such as Adaptive DBSCAN~\cite{ke2020} and trainable
Neural Networks~\cite{vinod1992}, which enhance accuracy and
robustness but typically require computational resources, and in some cases,
substantial training data.

The discussed approaches all attempt to address the limitations of the original
Hough transform, where noise introduces uncertainty in the Hough space, and
diffuse accumulations can result in missed detections or false positives.
However, each of these methods introduces additional parameters and, more
importantly, fails to resolve the fundamental issue: they detect only high peaks
rather than truly persistent ones in noisy environments.

\section{Topology-informed line detection}\label{sec:method}
As outlined in \Cref{subsec:ht}, the voting process for line detection in
Hough space $S$ is highly susceptible to noise in the sampled input image.
To address this, we introduce a topology-informed method for line detection in parameter
space $S$,
leveraging \ph.
This method eliminates the need for additional
parameters, while delivering more stable and reliable results.

The \ph-based method extends the original Hough transform by building upon the
point in the processing procedure where the accumulator array has already been
created.

Special attention must be given to the periodicity in Hough space and the defined observation limits.
Specifically, the neighborhood of $\theta=-\frac{\pi}{2}$ at the
beginning and $\theta=\frac{\pi}{2}$ at the end of the period in $S$ should be
considered glued together like a Möbius strip, as already discussed.

That is, the two edges of $S$, where $\theta = -\frac{\pi}{2}$ and
$\theta = \frac{\pi}{2}$ are connected in a diagonally opposing manner (to flip the sign of $\rho$).

The accumulator array derived from $S$ can be treated as an image, with
the voting counts representing the pixel values.
Applying the neighborhood definition outlined above, the \emph{superlevel-set}
filtration, as described in \Cref{subsec:ph}, is then performed on this array.
The resulting \pd reveals the most persistent lines in $S$:
lines with high persistence, indicating significance, are located further
from the \glspl{pd} diagonal, while lines resulting from noise appear
closer to the diagonal.

Given the \pd, the most persistent features are selected as detected lines.
As commonly done~\cite{edelsbrunner2008}, noisy topological features near the
diagonal, \ie, those with a persistence lower than a threshold $\nu$, are
filtered out.
The remaining features are then mapped back into $S$, where the
line parameters $\left( \rho, \theta \right)$ are determined for each of them.
Although the persistence threshold $\nu$ does not alter the sum of the
parameters, it simply replaces the previous threshold parameter.

In fact, within the~\pd we can illustrate both, the original and the \ph-based
voting mechanism, see \Cref{fig:ph-vs-ht}.
The original method simply performs a threshold at a specific \emph{birth} level
while we propose to perform a threshold at a specific persistence, leading to a
thresholding line parallel to the diagonal.

Additionally, the parameter type has shifted from being threshold-dependent on the input data
to an introduced parameter \(\nu\), which is independent of the input and can be
used in a ``one-size-fits-all'' setting.
A further key distinction between the original and the proposed method is that,
mathematically, our method tracks the
entire evolution of cubical complexes across the varying parameter, while
the original method simply constructs only a single cubical complex.

\begin{figure}
    \centering
    \resizebox{0.45\columnwidth}{!}
    {
        \begin{tikzpicture}[remember picture]
  \begin{axis}[
    ylabel={Death level},
    xlabel={Birth level},
    title={Persistence Diagram},
    width=10cm,
    height=10cm,
    axis equal image,
    xmin=0, xmax=250,
    ymin=0, ymax=250,
    x dir=reverse,
    y dir=reverse,
    scatter/classes={a={blue}},
    scatter src=explicit symbolic,
    legend pos=south east, % Position der Legende
    yticklabels={0,$-\infty$,50,100,150,200,250}
  ]

  % Punkte (Beispieldaten, bitte mit echten Werten ersetzen)
  \addplot[scatter,only marks,mark=*,black, forget plot,mark size=1pt] table {
    x y
    215 211
    213 182
    211 199
    209 170
    205 170
    203 198
    201 161
    199 175
    197 149
    195 183
    193 189
    191 190
    189 178
    187 140
    185 148
    183 164
    181 142
    177 127
    171 123
    167 162
    161 135
    159 142
    155 152
    153 145
    151 115
    147 140
    145 122
    143 109
    139 114
    137 114
    135 126
    131 127
    127 90
    123 101
    119 119
    115 76
    111 103
    107 101
    100 75
    85 60
    70 45
    55 40
    45 35
    35 33
  };

  % Linien
  \addplot[scatter,only marks,mark=*,olive,forget plot] table {
    x y
    230 5
    210 40
  };
  % Waagrechte Linie
  \addplot[mred,dashed, forget plot] coordinates {(200,0) (200,250)};
 
  \addplot[mblue,dashed, forget plot] coordinates {(255,180) (75,0)};
  %\addlegendimage{red,dashed}
  %\addlegendentry{baseline threshold} % Legende für die Linie
  % Diagonale Linie
  \addplot[black,dashed, forget plot] coordinates {(0,0) (250,250)};

  % **Manuelle Legende mit richtiger Farbe**
  \addlegendimage{only marks, mark=*, black}
  \addlegendentry{unwanted noise}

  \addlegendimage{only marks, mark=*, olive}
  \addlegendentry{detected lines}

  \addlegendimage{mblue,dashed}
  \addlegendentry{\ph-based method}

  \addlegendimage{mred,dashed}
  \addlegendentry{original method}

  \end{axis}
\end{tikzpicture}
    }
    \caption{Schematic comparison of line detection methods:
    fixed thresholding at a predefined value, as implemented in OpenCV, and
    adaptive thresholding based on feature persistence derived from \ph.}
    \label{fig:ph-vs-ht}
\end{figure}

The full advantages of applying \ph to the Hough space becomes evident upon
examining the \pd in \Cref{fig:ph-vs-ht}.
In the presence of noise near the \emph{peaks}, the Hough transform's threshold
parameter becomes highly sensitive, leading to false-positive line detections.
Adjusting this threshold requires a value-narrow trade-off between detecting all
\emph{relevant} lines and avoiding noise-induced artifacts.
For example, setting the threshold in this particular example low enough to exclude
noise also removes a desired line, while relaxing the threshold to retain both
intended lines results in detecting several irrelevant lines.
In contrast, the \ph-based approach is less sensitive to the chosen parameter
$\nu$.
The persistence of the lines we aim to detect is approximately equal to the
number of sampled points per line, while the persistence of noise-induced
artifacts corresponds to the expected noise level.

Furthermore, when a line is sampled with fewer points,
its corresponding point in the persistence diagram has a lower \emph{birth}
value.
In a strict thresholding regime, this means that as the number of points
decreases, a previously significant feature may fall below the threshold and go
undetected.
However, due to its diagonal orientation within the \pd, the proposed method
exhibits greater robustness to a decreasing sampling rate for different lines.

The computation of the \pd has a computational complexity of
$O\left(n \alpha\left(n\right) \right)$, while the filtration is
performed in $O\left(n\log\left(n\right)\right)$.
The storage requirements are linear, $O(n)$, where $n$ denotes the
number of complexes.

\section{Experiments and Evaluation}\label{sec:experiments-and-evaluation}
In the previous \Cref{sec:method} we showed on the~\pd, that
the original Hough transform has limitations concerning the threshold to be set.
From these observations, we derive the following claims:
(1) the original Hough transform is more sensitive to increasing noise levels
than the~\ph-based method, causing it to fail at lower noise levels, and
(2) at constant noise levels, a decreasing number of points per line compromises
the robustness of the original method more than that of the~\ph-based approach,
leading to earlier failure of the original method.
Additionally, leveraging \glspl{ph} mathematical properties, including
stability, we further assume that
(3) the proposed \ph-based method is Lipschitz stable.

We validate our claims with the following experiments\footnote{
    \ifthenelse{\boolean{anonymized}}{
        The URL to the Repository is removed to preserve anonymity for the double-blind
        review process.}{
        The source code for all the experiments is publicly available here: \url{https://github.com/JRC-ISIA/TopologicalHoughTransformation/}}
},
using generated images of the size $n_w \times n_h = 256 \times 256$
pixels.
In the first two experiments, each image contains two parallel
\emph{ground truth} lines with slope of $m=1$.
The y-intercepts are randomly sampled  as $b \in \left[ 50, \ldots, 100 \right]$
for the first line, while the second is symmetrically placed at $-b$.
Each image consists of a set of two-dimensional pixel coordinates
\( P = \left\{ (x_i, y_i) \mid 1 \leq i \leq n \right\} \), where $n$ is the
experiment dependent number of pixels uniformly sampled along
the line.
To introduce noise, each point $P_i$ is perturbed by a normally distributed shift orthogonal to the line,
$p_i^\prime \sim N\left(p_i, \frac{1}{\sqrt{2}} \varepsilon \id \right)$, where
$\varepsilon$ represents the
experiment-dependent noise level and $\id$ denotes the identity matrix.

As the baseline, we use the \emph{OpenCV} implementation of the Hough transform.
For both baseline and our proposed \ph-based method, the accumulator array is set to a
maximum resolution of $1$ pixel for the longest possible
$\rho$ and $1\deg$ for $\theta$, resulting in  $n_\rho=724$ and $n_\theta=180$.

The parameters -- threshold for the original Hough transform and the persistence
limit $\nu$ for the \ph-based method -- are optimized based on the F1-score,
balancing precision and recall.

For all experiments, we report accuracy, precision, and F1-score.
A detected line with parameter $\left( \rho^\prime, \theta^\prime \right)$
is considered correct if it falls within a band around the baseline, define as
the target line $\pm \varepsilon$.
Specifically, $\left( \rho^\prime, \theta^\prime \right)$ is deemed correct
if $\lvert \rho^\prime - \rho \rvert \leq \varepsilon $ and
$\lvert \theta^\prime - \theta \rvert \leq \frac{2\varepsilon}{n_w}$.

\begin{table}
    \centering
    \caption{Results of the first two experiments, as described in
        \Cref{subsec:exp-robust-noise,subsec:exp-sampling-lines}.}
    \label{tab:results}

    \begin{subtable}[h]{0.5\textwidth}
        \centering
        \subcaption{Comparison of metrics for both variants for $1500$ randomly
                 sampled images across varying noise level.}
        \label{tab:results_ex1}
        \begin{tabular}{r|c|c|c}
            \toprule
            \textbf{Method} & \textbf{Acc. (\%)} & \textbf{Prec. (\%)} & \textbf{F1 (\%)} \\
            \midrule
            \ph-based       & 47.78         & 91.39          & 64.66       \\
            original        & 24.67         & 24.73          & 39.57       \\
            \bottomrule
        \end{tabular}
    \end{subtable}
    \hfill
    \begin{subtable}[h]{0.45\textwidth}
        \centering
        \caption{Comparison of metrics for both variants
        with varying numbers of sampled points in the second line.}
        \label{tab:results_ex2}
        \begin{tabular}{c|c|c}
            \toprule
            \textbf{Acc. (\%)} & \textbf{Prec. (\%)} & \textbf{F1 (\%)} \\
            \midrule
            92.98              & 95.94               & 96.36  \\
            84.53              & 87.81              & 91.62  \\ 
            \bottomrule
        \end{tabular}
    \end{subtable}
\end{table}

\subsection{Robustness against varying noise}
\label{subsec:exp-robust-noise}

In the following experiment, we aim to validate our claim that the
proposed \ph-based method is more robust to noise than the original method.

To evaluate this, we compare both methods on in detecting
two parallel lines under varying noise level $\varepsilon$.
The lines are both parametrized as described earlier, with $n_1=150$ and $n_2=120$ sampled points,
respectively.
We generate a dataset of $100$ images, each containing
randomly sampled noise.
The noise levels $\varepsilon \in \left[ 5 \ldots 19 \right]$ are uniformly
distributed across to ensure a balanced representation of different
noise intensities.

The results, summarized in \Cref{tab:results_ex1}, show that our method
consistently outperforms the original approach across all evaluation metrics.

\Cref{fig:exp1_results_per_noise_level} provides a detailed analysis of the
results by noise level.
Both the methods exhibit similar performance trends
as noise increases.
However, the \ph-based method demonstrates greater resilience,
with its performance degrading more gradually.
At a noise level $\varepsilon = 10$, the original variant experiences a sharp drop in
approaching, whereas the \ph-based approach maintains a significant higher performance with
only a slight decline.

\iffalse
Info für Matrin: Ex. 1:
Normalverteilter Noise, standartabw. = 1..15, µ= 0
Linie wird ok gewertet wenn rho und theta im Rechteck 1 Standartabweichung + Rundungstoleranz (2* runden, daher 2Pixel) sind
Parallele Linien, Winkel 45°
Beide Linien haben in y-Richtung 50-100 Pixel Abstand zur Bildmitte
Threshold baseline so optimiert dass zumindest anfangs beide Linien erkannt werden
Linien haben 400 und 350 Pixel
\fi

\begin{figure}
    \centering
    \begin{subfigure}{0.48\linewidth}
        \centering
        \begin{tikzpicture}
            \begin{axis}[
                width=\columnwidth,
                height=4cm,
                xlabel={Noiselevel $\varepsilon$},
                xmin=5, xmax=19,
                ymin=0, ymax=1.1,
                %legend pos=north east,
                %legend style={font=\small},
                grid=both,
                major grid style={line width=.2pt, draw=gray!50},
                minor grid style={line width=.1pt, draw=gray!20},
            ]
            % Accuracy data baseline
            \addplot[
                color=mred,
                mark=*,
                thick
            ] coordinates {
                (5, 0.961538461538462) (6, 0.98) (7, 0.8) (8, 0.525) (9, 0.275) (10, 0.09) (11, 0.025) (12, 0.005) (13, 0.01) (14, 0) (15, 0) (16, 0) (17, 0) (18, 0) (19, 0)
            };
            %\addlegendentry{baseline}

            % Accuracy data out method
            \addplot[
                color=mblue,
                mark=square*,
                thick
            ] coordinates {
                (5, 0.975369458128079) (6, 0.970443349753695) (7, 0.970731707317073) (8, 0.951456310679612) (9, 0.952153110047847) (10, 0.800847457627119) (11, 0.729838709677419) (12, 0.702602230483271) (13, 0.575949367088608) (14, 0.46684350132626) (15, 0.422110552763819) (16, 0.31588785046729) (17, 0.262910798122066) (18, 0.228378378378378) (19, 0.168421052631579)
            };
            %\addlegendentry{our method}

            \end{axis}
        \end{tikzpicture}
        \caption{Accuracy}
        \label{fig:exp1-accuracy}
    \end{subfigure}%
    %\hfill
    \begin{subfigure}{0.48\linewidth}
        \centering
        \begin{tikzpicture}
            \begin{axis}[
                width=\columnwidth,
                height=4cm,
                xlabel={Noiselevel $\varepsilon$},
                yticklabel=\empty,
                xmin=5, xmax=19,
                ymin=0, ymax=1.1,
                %legend pos=north east,
                %legend style={font=\small},
                grid=both,
                major grid style={line width=.2pt, draw=gray!50},
                minor grid style={line width=.1pt, draw=gray!20},
            ]
            % Precision data baseline
            \addplot[
                color=mred,
                mark=*,
                thick
            ] coordinates {
                (5, 1) (6, 0.98) (7, 0.8) (8, 0.525) (9, 0.275) (10, 0.09) (11, 0.025) (12, 0.005) (13, 0.01) (14, 0) (15, 0) (16, 0) (17, 0) (18, 0) (19, 0)
            };
            %\addlegendentry{baseline}

            % Precision data our method
            \addplot[
                color=mblue,
                mark=square*,
                thick
            ] coordinates {
                (5, 1) (6, 0.985) (7, 0.995) (8, 0.98) (9, 0.995) (10, 0.945) (11, 0.905) (12, 0.945) (13, 0.91) (14, 0.88) (15, 0.84) (16, 0.845) (17, 0.84) (18, 0.845) (19, 0.8)
            };
            %\addlegendentry{our method}

            \end{axis}
        \end{tikzpicture}
        \caption{Precision}
        \label{fig:exp1-precision}
    \end{subfigure}
    \caption{The results of the experiment targeting our first stated claim, comparing the Accuracy (\cref{fig:exp1-accuracy}) and (\cref{fig:exp1-precision}) of the orignial-
        ({\protect\tikz \protect\draw[mred, mark=*, mark options={scale=1.2}] plot coordinates {(0,0)};})
        compared with our proposed \ph-based method
        ({\protect\tikz \protect\draw[mblue, mark=square*, mark options={scale=1.2}] plot coordinates {(0,0)};}).
    The constant performance of both methods with noise levels $\varepsilon < 5$ are ommitted for brevity.}
    \label{fig:exp1_results_per_noise_level}
\end{figure}

\begin{figure}
    \centering
    \includegraphics[width=0.99\textwidth]{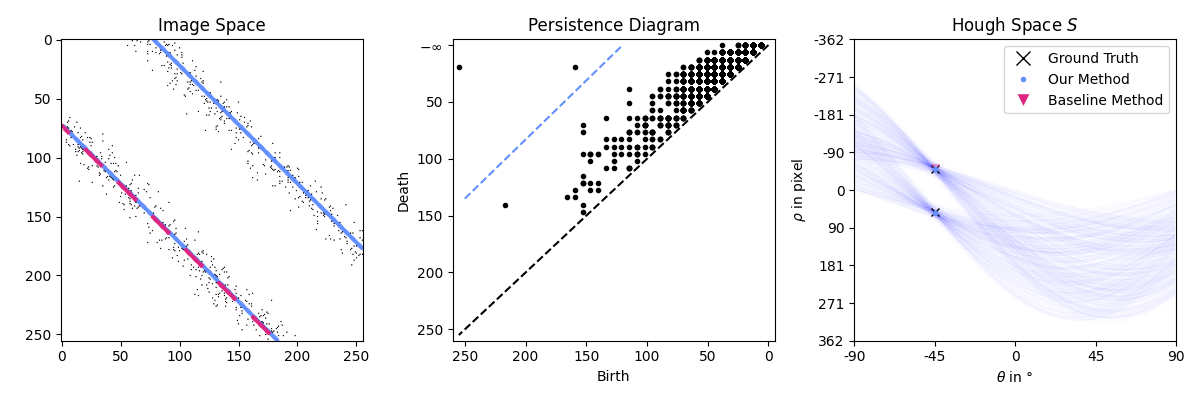}
    \caption{
    Result for the experiment with the noised, parrallel lines.
    The first plot shows the sampled points and the one line detected by
    the original ({\protect\tikz \protect\draw[mred, thick] (-0.15,0.12) -- (0.15,0.12) node[draw=none] at (0,0.15) {}  node[draw=none] at (0,0.2) {} ;}) and  both lines detected by the \ph-based ({\protect\tikz \protect\draw[mblue, thick] (-0.15,0.12) -- (0.15,0.12) node[draw=none] at (0,0.15) {}  node[draw=none] at (0,0.2) {} ;}) method.
    The second plot visualizes the \pd, highlighting the applied threshold $\nu$.
    The third plot depicts the Hough space $S$, illustrating the sinusoidal
    curves and the corresponding detected lines.}
    \label{fig:result_ex1_single_run}
\end{figure}

\Cref{fig:result_ex1_single_run} provides a detailed comparison of both methods on a specific
example with a noise level of $\varepsilon = 8$.
In the image space, our proposed method successfully detects both lines, whereas the original method
correctly identifies only one.
The underlying reason becomes evident when analyzing the \pd:
while the \ph-based approach detects lines based on the persistence of topological
features, the original method effectively \emph{cuts} the \pd vertically along the
\emph{birth} axis.
As discussed in \Cref{sec:method}, this threshold-based approach is highly sensitive to parameter
selection and heavily depends on the dataset.
The visualization of the Hough space $S$ further illustrates the detected
lines by highlighting their corresponding intersections along the sinusoidal curves.

\subsection{Robustness against irregular number of samples per lines}
\label{subsec:exp-sampling-lines}

In this experiment, we validate our second claim that the \ph-based method
is less sensitive to variations in the number of sampled points per line.
To evaluate this, we follow a setup similar to the previous experiment:
we use two parallel lines, parametrized as before, with a fixed noise level of
$\varepsilon=3$.
The first line is always sampled with $n_1=500$ points, while the second line has
$n_2 \in \left[ 150 \ldots 500 \right]$ in steps of $50$.
For each value of $n_2$, we generate $10$ images.

The overall performance across the dataset is summarized in
\Cref{tab:results_ex2}.
As in the previous experiment, our proposed method consistently outperforms the
baseline across all evaluation metrics.

\Cref{fig:exp2_results_per_num_points} provides a detailed breakdown of performance across
varying values of $n_2$.
It is evident that the original method fails earlier as the number of
sampled points in the second line decreases.
While the \ph-based method also experiences a decline in performance,
it maintains a clear advantage over the original variant, exhibiting a much
slower degradation.

\iffalse
Info für Matrin: Ex. 1:

\fi

\begin{figure}
    \centering
    \begin{subfigure}{0.48\linewidth}
        \centering
        \begin{tikzpicture}
            \begin{axis}[
                width=\linewidth,
                height=4cm,
                xlabel={Pointcount $n_2$},
                xmin=50, xmax=500,
                ymin=0.4, ymax=1.1,
              %  legend pos=south east,
              %  legend style={font=\small},
                grid=both,
                major grid style={line width=.2pt, draw=gray!50},
                minor grid style={line width=.1pt, draw=gray!20},
            ]
            % Accuracy data baseline
            \addplot[
                color=mred,
                mark=*,
                thick
            ] coordinates {
                (500, 0.930232558139535) (450, 0.930232558139535) (400, 0.956937799043062) (350, 0.975609756097561) (300, 0.961538461538462) (250, 0.861244019138756) (200, 0.565) (150, 0.557213930348259) (100, 0.514705882352941) (50, 0.490196078431373)
            };
            %\addlegendentry{baseline}

            % Accuracy data out method
            \addplot[
                color=mblue,
                mark=square*,
                thick
            ] coordinates {
                (500, 0.980392156862745) (450, 0.975609756097561) (400, 0.956937799043062) (350, 0.956937799043062) (300, 0.952380952380952) (250, 0.951456310679612) (200, 0.887254901960784) (150, 0.775609756097561) (100, 0.57487922705314) (50, 0.497512437810945)
            };
            %\addlegendentry{our method}

            \end{axis}
        \end{tikzpicture}
        \caption{Accuracy}
        \label{fig:exp2-accuracy}
    \end{subfigure}%
    \hfill
    \begin{subfigure}{0.48\linewidth}
        \centering
        \begin{tikzpicture}
            \begin{axis}[
                width=\columnwidth,
                height=4cm,
                xlabel={Pointcount $n_2$},
                yticklabel=\empty,
                xmin=50, xmax=500,
                ymin=0.4, ymax=1.1,
               % legend pos=south east,
               % legend style={font=\small},
                grid=both,
                major grid style={line width=.2pt, draw=gray!50},
                minor grid style={line width=.1pt, draw=gray!20},
            ]
            % Precision data baseline
            \addplot[
                color=mred,
                mark=*,
                thick
            ] coordinates {
                (500, 1) (450, 1) (400, 1) (350, 1) (300, 1) (250, 0.9) (200, 0.565) (150, 0.56) (100, 0.525) (50, 0.5)
            };
            %\addlegendentry{baseline}

            % Precision data our methdo
            \addplot[
                color=mblue,
                mark=square*,
                thick
            ] coordinates {
                (500, 1) (450, 1) (400, 1) (350, 1) (300, 0.995024875621891) (250, 0.98) (200, 0.905) (150, 0.795) (100, 0.586206896551724) (50, 0.5)
            };
            %\addlegendentry{our method}

            \end{axis}
        \end{tikzpicture}
        \caption{Precision}
        \label{fig:exp2-precision}
    \end{subfigure}
    \caption{The results of the experiment targeting our second stated claim, comparing the Accuracy (\cref{fig:exp2-accuracy})
        and (\cref{fig:exp2-precision}) of the orignial-
        ({\protect\tikz \protect\draw[mred, mark=*, mark options={scale=1.2}] plot coordinates {(0,0)};})
        compared with our proposed \ph-based method
        ({\protect\tikz \protect\draw[mblue, mark=square*, mark options={scale=1.2}] plot coordinates {(0,0)};}).
    }
    \label{fig:exp2_results_per_num_points}
\end{figure}

\begin{figure}
    \centering
    \includegraphics[width=0.99\textwidth]{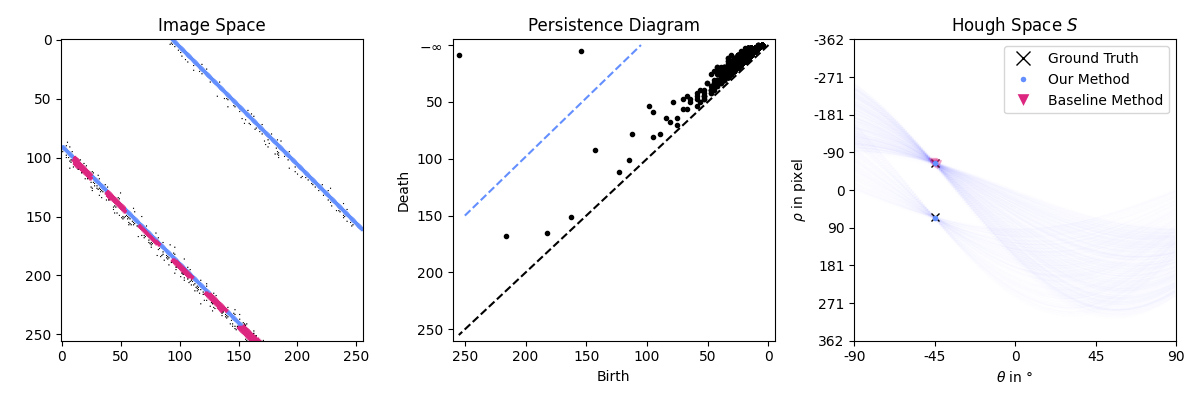}
    \caption{Result for the experiment with the unevenly sampled, parrallel lines.
    The first plot shows the sampled points and the one line detected by
    the original ({\protect\tikz \protect\draw[red, thick] (-0.15,0.12) -- (0.15,0.12) node[draw=none] at (0,0.15) {}  node[draw=none] at (0,0.2) {} ;}) and  both lines detected by the \ph-based ({\protect\tikz \protect\draw[mblue, thick] (-0.15,0.12) -- (0.15,0.12) node[draw=none] at (0,0.15) {}  node[draw=none] at (0,0.2) {} ;}) method.
    The second plot visualizes the \pd, highlighting the applied threshold $\nu$.
    The third plot depicts the Hough space $S$, illustrating the sinusoidal curves and the corresponding detected
    lines.}
    \label{fig:result_ex2_single_run}
\end{figure}

\Cref{fig:result_ex2_single_run} provides a detailed comparison of both methods for a specific
case with $n_2=400$.
In the image space, our method successfully detects both lines, whereas
the original variant only detects one.
The \pd clearly illustrates the reason for the missed detection:
the threshold of the original method is set to detect one line, but detecting
the second line -- given its lower number of sampled points -- would require
lowering this threshold.
However, doing so would also introduce two false positive near the first line.
In contrast, the \ph-based approach effectively distinguishes true lines from noise.

\subsection{Empirical Stability}\label{subsec:exp-stability}

With our third claim, we argue that the proposed \ph-based method exhibits
stability properties in the sense of Lipschitz continuity.
While we do not attempt to formally prove Lipschitz stability in this work,
we provide empirical evidence.

\begin{definition}[Lipschitz Continouity]\label{def:lipshitz}
    A function \( f: X \to Y \) mapping from a metric space \( (X, d_X) \) to
    a metric space \( (Y, d_Y) \) is said to be \emph{Lipschitz continuous} if there is a
    constant \( L \geq 0 \) such that
    \(
        d_Y(f(x_1), f(x_2)) \leq L \cdot d_X(x_1, x_2).
    \)
    for all \(  x_1, x_2 \in X \).
\end{definition}

Lipschitz continuity bounds the effect of input perturbations on the output.
A small perturbation in the input leads to a correspondingly small change in the
output. It is therefore a notion for stability.

Given our mapping from image space via Hough space to \glspl{pd},
\(
    P \xrightarrow{f_1} S \xrightarrow{f_2} PD,
\)
our aim is to demonstrate that perturbations in the image
space, propagated through $f_1 \circ f_2$ remain bounded by some constant $L$ within the \glspl{pd}.

\begin{remark}[Stability of \ph]\label{remark:stability-ph}
    The mapping of $f_2$ is already proven to be
    Lipschitz stable~\cite{cohen-steiner2007}.
    Therefore, within the chain of mappings, the stability properties of $f_1$
    remain underexplored and are the focus of this experiment.
\end{remark}

For this setup, we use a single line parallel to the x-axis with $n=50$, slope $m=0$,
y-intercept $b=120$, positioned near the center of the image.
In each of $35$ iterations, one of the $50$ pixel $p_i = (x_i, y_i)$ is randomly selected (without replacement) and perturbed, generating a new image where
$y_i^\prime \sim N\left(y_i, 10 \right)$.
This process iteratively transforms the image from a straight line to a noisy version.
The experiment is repeated $10$ times to evaluate different initial setups.
For each iteration, the difference between the original, non-perturbed and the perturbed image is quantified
using the $1$-Wasserstein distance $d_W$,
while the \emph{Bottleneck distance} $d_B$ is applied to the corresponding \glspl{pd}.

\input{tikz/exp3_results}

The results are show in \Cref{fig:result_exp3-stability-histogram}, where the
bottleneck distance $d_B$ of the \glspl{pd} is plotted over the corresponding
$1$-Wasserstein distance in image space.
In \Cref{fig:exp3-stability}, we observe a clear constraint that is
never exceeded in the relationship between distance metrics from input to output.
The bottleneck distance $d_B$ converges toward $25$ due to the choice of $n=50$
and the presence of a high noise regime.
Instead of a single significant peak, multiple low-count peaks appear near the
diagonal.
As a result, the maximal shift distance between two points determines the
observed convergence.
\Cref{fig:exp3-ratio} illustrates the ratio of the bottleneck distance $d_B$ of
the \glspl{pd} to the Wasserstein distance $d_W$ of the input.
This ratio, derived as a reformulation of \Cref{def:lipshitz} in terms of $L$,
ensures that $L \geq 1$, thereby showing Lipschitz continuity in this
specific experimental setting.
However, when changing the metric on the input images to the bottleneck distance,
stability guarantees are no longer provided.

\section{Discussion \& Conclusion}\label{sec:discussion-conclusion}

The experimental results strongly support the claims in
\Cref{sec:experiments-and-evaluation} on our experimental setups.
We demonstrate that increasing noise has less impact on our method
compared to the original line detection method.
While the original method's performance rapidly degrades, our approach remains
more robust.
Additionally, we show that when the number of points on a line decreases while
the number of points on another line remains constant, our method remains
resilient.

Our third claim addresses the Lipschitz stability of our proposed method.
The experiment clearly shows stability when using the $1$-Wasserstein distance
in the input space.
However, the stability argument is not satisfied when using the bottleneck
distance between the input images.
Achieving stability using the bottleneck distance requires further research.

Despite the compelling results, the proposed \ph-based method also has limitations.
Computational complexity for line detection increases over to the original
Hough transform, as computing the filtration introduces an additional
$O\left(n\log\left(n\right)\right)$ complexity, while constructing the \pd adds
$O(n \alpha(n))$.
Moreover, while our experimental  results suggest that our method is Lipschitz
stable, we lack formal theoretical evidence to support this claim.
Yet, this conclusion is limited to the $1$-Wasserstein distance metric in the
input domain.

Building on this work, we identify a wide range of promising research
directions.

A natural extension is to adapt our method for detecting shapes beyond
lines, such as curves, circles, and more generally, arbitrary template
shapes using the Generalized Hough transform.

Additionally, the integration of the \ph-based method with other variants of the original Hough transform,
such as the Kernel Hough transform, or
the \emph{random sampling} method,
presents another compelling avenue for further exploration.

For the proposed \ph-based method, we do not provide a formal theoretical proof
of its general stability.
However, as demonstrated in \Cref{subsec:exp-stability}, our method exhibits
empirical stability within the given setting.
Nonetheless, we emphasize the importance of conducting a comprehensive stability
analysis and developing mechanisms to further enhance its robustness.

In this work, we introduced \ph-based method for line detection, a variant of the original Hough transform, that
leverages topological structure of the Hough space for more robust line detection.
Through our experiments, we demonstrate that the \ph-based method outperforms the original Hough transform, exhibiting
greater robustness to varying noise levels
and different numbers of points per line
in the input image.
In both experiments, our approach achieved higher
detection performance, validating our initial claims.

Beyond its empirical advantages, this work also initiates a discussion on the
mathematical stability properties of the Hough transform, a topic that has received little
attention.
Establishing formal guarantees on stability could further enhance the
reliability of the Hough transform in real-world applications.

Moreover, we hope this work will inspire researchers from diverse fields,
where \ph and \tda are underrepresented, to explore the potential of these
tools in their methods.
We firmly believe that applying \tda techniques represents a valuable
opportunity to generate novel insights across a range of problems.
Even those problems considered ``solved'' may have untapped potential and
can benefit from re-examination and further exploration.
We hope this work encourages further exploration of \tda in computer vision and
beyond.

\begin{credits}
    \ifthenelse{\boolean{anonymized}}
    {
        \subsubsection{\ackname} Lorem ipsum dolor sit amet, consectetur
        adipiscing elit. Sed consequat mauris et pellentesque vestibulum.
        Aenean convallis, urna sed ornare lobortis, turpis arcu rhoncus tellus,
        sit amet tincidunt odio ipsum in.
    }{
        \subsubsection{\ackname} The financial support by the Christian Doppler Research Association, the
        Austrian Federal Ministry for Digital and Economic Affairs and the Federal State
        of Salzburg is gratefully acknowledged.
    }
\end{credits}

\bibliographystyle{splncs04}
\bibliography{./topo-lines-paper}

\begin{thebibliography}{10}
\providecommand{\url}[1]{\texttt{#1}}
\providecommand{\urlprefix}{URL }
\providecommand{\doi}[1]{https://doi.org/#1}

\bibitem{ballard1981}
Ballard, D.: Generalizing the {{Hough}} transform to detect arbitrary shapes.
  Pattern Recognition  \textbf{13}(2),  111--122 (1981)

\bibitem{opencv_library}
Bradski, G.: {The OpenCV Library}. Dr. Dobb's Journal of Software Tools  (2000)

\bibitem{cohen-steiner2007}
{Cohen-Steiner}, D., Edelsbrunner, H., Harer, J.: Stability of {{Persistence
  Diagrams}}. Discrete \& Computational Geometry  \textbf{37}(1),  103--120
  (Jan 2007)

\bibitem{du2012}
Du, S., Tu, C., van Wyk, B.J., Ochola, E.O., Chen, Z.: Measuring {{Straight
  Line Segments Using HT Butterflies}}. PLOS ONE  \textbf{7}(3),  e33790 (2012)

\bibitem{duda1972}
Duda, R.O., Hart, P.E.: Use of the {{Hough}} transformation to detect lines and
  curves in pictures. Communications of the ACM  \textbf{15}(1),  11--15 (1972)

\bibitem{edelsbrunner2008}
Edelsbrunner, H., Harer, J.: Persistent homology—a survey. In: Contemporary
  {{Mathematics}}, vol.~453, pp. 257--282. American Mathematical Society (2008)

\bibitem{fernandes2008}
Fernandes, L.A., Oliveira, M.M.: Real-time line detection through an improved
  {{Hough}} transform voting scheme. Pattern Recognition  \textbf{41}(1),
  299--314 (2008)

\bibitem{hassanein2015}
Hassanein, A.S., Mohammad, S., Sameer, M., Ragab, M.E.: A {{Survey}} on {{Hough
  Transform}}, {{Theory}}, {{Techniques}} and {{Applications}}  (2015)

\bibitem{Huber2020}
Huber, S.: Persistent homology in data science. In: Proc.\ 3rd Int.\ Data
  Science Conference (iDSC'20). Data Science -- Analytics and Applications,
  Dornbirn, Austria (virtual) (May 2020)

\bibitem{ke2020}
Ke, R., Feng, S., Cui, Z., Wang, Y.: Advanced framework for microscopic and
  lane-level macroscopic traffic parameters estimation from {{UAV}} video. IET
  Intelligent Transport Systems  \textbf{14}(7),  724--734 (2020)

\bibitem{mukhopadhyay2015}
Mukhopadhyay, P., Chaudhuri, B.B.: A survey of {{Hough Transform}}. Pattern
  Recognition  \textbf{48}(3),  993--1010 (Mar 2015)

\bibitem{stephens1991}
Stephens, R.S.: Probabilistic approach to the {{Hough}} transform. Image and
  Vision Computing  \textbf{9}(1),  66--71 (1991)

\bibitem{vinod1992}
Vinod, V., Chaudhury, S., Ghose, S., Mukherjee, J.: A connectionist approach
  for peak detection in {{Hough}} space. Pattern Recognition  \textbf{25}(10),
  1253--1264 (1992)

\bibitem{xu2014a}
Xu, Z., Shin, B.S.: Line {{Segment Detection}} with {{Hough Transform Based}}
  on {{Minimum Entropy}}. In: Image and {{Video Technology}}. pp. 254--264.
  Springer (2014)

\bibitem{xu2014}
Xu, Z., Shin, B.S.: A {{Statistical Method}} for {{Peak Localization}} in
  {{Hough Space}} by {{Analysing Butterflies}}. In: Image and {{Video
  Technology}}. pp. 111--123. Springer (2014)

\bibitem{xu2015}
Xu, Z., Shin, B.S., Klette, R.: A statistical method for line segment
  detection. Computer Vision and Image Understanding  \textbf{138},  61--73
  (2015)

\end{thebibliography}

\end{document}